%% file: root.tex
\documentclass[letterpaper, 10pt, journal, twoside]{IEEEtran}
\usepackage[utf8]{inputenc}

\usepackage{amsmath}
\usepackage{bm}
\usepackage{nicefrac}
\usepackage{xcolor}
\usepackage{graphicx}
\usepackage[binary-units=true]{siunitx}
\usepackage{multirow}
\usepackage{hyperref}

\newcommand{\figref}[1]{Fig.~\ref{#1}}
\newcommand{\tabref}[1]{Table~\ref{#1}}

\newcommand{\etal}{\emph{et~al.}}

\newcommand{\rf}[1]{\scriptscriptstyle \mathtt{#1}}
\newcommand{\distributed}{\!\!\sim\!\!}
\newcommand{\REAL}{\mathrm{I\!R}}
\newcommand{\norm}[1]{\left\lVert#1\right\rVert}

\DeclareMathOperator{\Exp}{Exp}
\DeclareMathOperator{\Log}{Log}
\DeclareMathOperator*{\argmax}{arg\,max}
\DeclareMathOperator*{\argmin}{arg\,min}
\DeclareMathOperator{\tr}{tr}

\include{tables}

\begin{document}

\title{An Analytical Solution to the IMU Initialization Problem for Visual-Inertial Systems}

\author{David Zuñiga-Noël, Francisco-Angel Moreno, Javier Gonzalez-Jimenez}

\maketitle

\begin{abstract}
The fusion of visual and inertial measurements is becoming more and more popular in the robotics community since both sources of information complement well each other. However, in order to perform this fusion, the biases of the Inertial Measurement Unit (IMU) as well as the direction of gravity must be initialized first. Additionally, in case of a monocular camera, the metric scale is also needed.
The most popular visual-inertial initialization approaches rely on accurate vision-only motion estimates to build a non-linear optimization problem that solves for these parameters in an iterative way. In this paper, we rely on the previous work in~\cite{mur2017} and propose an analytical solution to estimate the accelerometer bias, the direction of gravity and the scale factor in a maximum-likelihood framework. This formulation results in a very efficient estimation approach and, due to the non-iterative nature of the solution, avoids the intrinsic issues of previous iterative solutions.
We present an extensive validation of the proposed IMU initialization approach and a performance comparison against the state-of-the-art approach described in~\cite{campos2020} with real data from the publicly available EuRoC dataset, achieving comparable accuracy at a fraction of its computational cost and without requiring an initial guess for the scale factor.
We also provide a C++ open source reference implementation.

\end{abstract}

\section{Introduction}
Visual-Inertial (VI) systems are ubiquitous in robotics nowadays, mainly because visual and inertial sensors are usually inexpensive, have low-power requirements, and provide information complementary to each other. For instance, a monocular camera is an exteroceptive sensor that allows recovering the geometry of the environment up to a scale factor, while an Inertial Measurement Unit (IMU) is a proprioceptive sensor that allows rendering the metric scale observable and provides robust frame-to-frame motion priors. Consequently, VI systems have become minimal set-ups for metric visual SLAM and thus have found applications ranging from AR/VR to autonomous navigation, both aerial and ground~\cite{forster2017}.

However, before visual and inertial measurements can be fused, some parameters of the IMU need to be initialized, namely: the gravity direction, and the gyroscope and accelerometer biases. Not only that, these biases slowly change over time and the gravity direction depends on the initial pose, so they have to be estimated online. Additionally, knowing the metric scale is also required for VI fusion in case of a monocular system. The initialization of these parameters is critical for the performance of the VI system, and has to be fast and accurate, otherwise the fusion will be delayed and/or convergence issues might appear~\cite{campos2020}.

Two main approaches to VI initialization can be found in the literature: (i) solving jointly the visual Structure-from-Motion (SfM) problem and the IMU initialization~\cite{martinelli2014, kaiser2017} or (ii) first addressing SfM problem only and then initializing the IMU from the estimate motion~\cite{mur2017, qin2017}.
The former assumes that the camera poses can be approximated from inertial measurements and requires the same visual features to be detected across all frames used for initialization. This hinders the practical application of the method, which additionally suffers form numerical stability problems. 
On the other hand, the latter assumes that the vision-only motion estimation is very accurate so that the IMU parameters can be estimated from them. In case the reader is wondering, this is actually the case for state-of-the-art monocular SLAM systems, such as ORB\_SLAM~\cite{orb_slam}, which achieves a localization accuracy in the order of a few centimeters for room-scaled environments. Following this direction, Campos \etal~\cite{campos2020} recently proposed to solve the involved non-linear optimization problem in a maximum-a-posteriori (MAP) framework, resulting in an accurate and efficient approach. However, due to the iterative nature of the solution, a good initial guess for all parameters is required beforehand. This can be done for the direction of gravity, and the IMU biases can be safely initialized to zero, but estimating the scale factor remains a problem since it cannot be known \emph{a priori} for a monocular vision system.

In this paper, we rely on the previous work of Mur-Artal and Tardós~\cite{mur2017} and propose a new solution to the visual-inertial initialization problem within a maximum likelihood estimation (MLE) framework. The bias for the gyroscope is estimated iteratively, as in~\cite{mur2017}, but our main contribution in this paper lies in an analytic solution for the rest of the parameters (i.e. scale, gravity direction and accelerometer bias).
The proposed solution is first validated and then evaluated with real data from the EuRoC~\cite{euroc} dataset and compared to the recent work of Campos \etal~\cite{campos2020}, showing state-of-the-art accuracy while being more efficient (about \num{7} times faster) and without requiring an initial guess for the scale factor.
We also released our implementation for the benefit of the community.

\section{Related Work}
Martinelli proposed in~\cite{martinelli2014} a closed-form solution to the joint VI-SfM problem. His approach assumes that a set of 3D points can be observed in all frames during initialization and that the IMU measurements can be used to approximate the camera poses. The scale, gravity direction, accelerometer bias, and initial velocity are solved along with the depth of the visual features by building a system of linear equations. Kaiser \etal~extended this approach in~\cite{kaiser2017} to also take into account the gravity magnitude and estimate the gyroscope bias, resulting in a non-linear optimization problem solved iteratively. However, the main issue with this approach is that the 3D error of the visual features is minimized instead of the reprojection error, which results in large unpredictable errors as shown in~\cite{campos2019}.

Modern visual odometry and SLAM systems achieve higher accuracy than IMU integration through Bundle-Adjustment (BA). This inspired Mur-Artal and Tardós~\cite{mur2017} to handle the IMU and visual initialization separately. Once the vision-based subsystem is initialized and a first visual BA is performed, they consider the initialization of the IMU parameters. Qin and Shen presented a similar approach in~\cite{qin2017} but leaving the estimation of the accelerometer bias to the VI-BA. In both cases, the IMU initialization problem is formulated independently for the gyroscope and the accelerometer as two ordinary least-squares problems. Note that, from a MAP perspective, ignoring the measurement uncertainties yields suboptimal results.

Campos \etal~addressed these issues in~\cite{campos2020} and formulated the IMU initialization problem within a MAP framework. However, the resulting non-linear optimization problem is solved iteratively, and thus requires a proper initial guess of the parameters. In fact, they perform three optimizations with different initial scales in parallel to successfully bootstrap the unknown initial scale factor. In our work, though, we formulate the IMU initialization problem as two independent problems similar to \cite{mur2017, qin2017} but within a MLE framework. Additionally, we propose an analytical solution for the accelerometer subproblem, thus removing the need of finding iteratively the scale factor and gravity direction, consequently achieving a much faster initialization.

\section{Problem Formulation}
\subsection{IMU measurement model}
First, we briefly describe in this section the considered IMU measurement model. For a detailed description and derivations, we refer the interested reader to~\cite{kok2017}. An IMU measures both the angular velocity and the specific forces in the body frame with respect to an inertial frame. Ignoring the effects of Earth's rotation, the simplified IMU measurement model can be defined as:
\begin{align}
    \label{eq:gyro_model}
    {}_{\rf{B}}\tilde{\bm{\omega}}(t) &= {}_{\rf{B}}\bm{\omega}(t) + \bm{b}^g + \bm{\eta}^g\\
    \label{eq:acc_model}
    {}_{\rf{B}}\tilde{\bm{a}}(t) &= \mathtt{R}^\top_{\rf{WB}}(t)({}_{\rf{W}}\bm{a}(t) - {}_{\rf{W}}\bm{g}) +  \bm{b}^a + \bm{\eta}^a,
\end{align}
in terms of the instantaneous angular velocity ${}_{\rf{B}}\bm{\omega} \in \REAL^3$ and the linear acceleration ${}_{\rf{W}}\bm{a} \in \REAL^3$ of the sensor. In this notation, we use prefixes to denote in which reference frame the corresponding elements are expressed. For instance, \texttt{W} and \texttt{B} refer to the world and body frames, respectively. The biases are represented by $\bm{b}^g, \bm{b}^a \in \REAL^3$ for the gyroscope and accelerometer respectively, while ${}_{\rf{W}}\bm{g} \in \REAL^3$ stands for the gravity vector. In turn, the pose of the IMU is described by the rigid body transformation $\{\mathtt{R}_{\rf{WB}}, {}_{\rf{W}}\bm{p}\} \in \text{SE(3)}$, which includes both the rotation and the position, mapping the sensor frame \texttt{B} coordinates to \texttt{W}. Finally, the gyroscope and accelerometer measurements are corrupted by zero-mean Gaussian noise: $\bm{\eta}^g\distributed\mathcal{N}(\bm{0}, \mathbf{\Sigma}^g),\ \bm{\eta}^a\distributed\mathcal{N}(\bm{0}, \mathbf{\Sigma}^a)$, respectively. Note that the biases are considered fixed in this model, since the IMU initialization will only last for a short period of time  (typically 2-10s).

Applying Euler integration, the pose and instantaneous velocity of the sensor ${}_{\rf{W}}\bm{v} \in \REAL^3$ can be expressed in terms of the IMU measurements in discrete time intervals as:
\begin{align}
    \label{eq:rotation_integral}
    \mathtt{R}_{\rf{WB}}(t + \Delta t) &=\mathtt{R}_{\rf{WB}}(t)\Exp\big(({}_{\rf{B}}\tilde{\bm{\omega}}(t) - \bm{b}^g - \bm{\eta}^{gd})\Delta t\big)\\
    \label{eq:velocity_integral}
    \begin{split}
        {}_{\rf{W}}\bm{v}(t + \Delta t) &= {}_{\rf{W}}\bm{v}(t) + {}_{\rf{W}}\bm{g}\Delta t\\                &+\mathtt{R}_{\rf{WB}}(t)({}_{\rf{B}}\tilde{\bm{a}}(t) - \bm{b}^a - \bm{\eta}^{ad})\Delta t
    \end{split}\\
    \label{eq:position_integral}
    \begin{split}
        {}_{\rf{W}}\bm{p}(t + \Delta t) &= {}_{\rf{W}}\bm{p}(t) + {}_{\rf{W}}\bm{v}(t)\Delta t + \frac{1}{2}{}_{\rf{W}}\bm{g}\Delta t^2\\
        &+ \frac{1}{2}\mathtt{R}_{\rf{WB}}(t)({}_{\rf{B}}\tilde{\bm{a}}(t) - \bm{b}^a - \bm{\eta}^{ad})\Delta t^2
    \end{split}
\end{align}
were $\Delta t$ represents the integration time and $\Exp(\cdot)$ is the vector-valued exponential map of SO(3). The covariance of the discrete-time noise for the gyroscope and accelerometer are $\mathbf{\Sigma}^{gd} = \nicefrac{1}{\Delta t}\mathbf{\Sigma}^g$ and $\mathbf{\Sigma}^{ad} = \nicefrac{1}{\Delta t}\mathbf{\Sigma}^a$, respectively~\cite{trawny2005}.

We adopt the formulation in~\cite{forster2017} to preintegrate the IMU measurements from keyframe to keyframe. In this way, the poses and the velocity of the IMU at the $i$-th and $j$-th time steps are related by:
\begin{align}
    \label{eq:preintegrated_rotation}
    \begin{split}
        \Delta \mathtt{R}_{ij}\ &\dot{=}\ \mathtt{R}_{i}^\top\mathtt{R}_{j}\\
        &= \prod_{k=i}^{j-1} \Exp\big((\tilde{\bm{\omega}}_k - \bm{b}^g - \bm{\eta}^{gd})\Delta t\big)
    \end{split}\\
    \label{eq:preintegrated_velocity}
    \begin{split}
        \Delta \bm{v}_{ij}\ &\dot{=}\ \mathtt{R}_{i}^\top\big(\bm{v}_j - \bm{v}_i - \bm{g}\Delta t_{ij} \big)\\
        &= \sum_{k=i}^{j-1} \Delta \mathtt{R}_{ik} (\tilde{\bm{a}}_k - \bm{b}^a - \bm{\eta}^{ad})\Delta t
    \end{split}\\
    \label{eq:preintegrated_position}
    \begin{split}
        \Delta \bm{p}_{ij} \ &\dot{=}\ \mathtt{R}_{i}^\top\big(\bm{p}_j - \bm{p}_i - \bm{v}_i\Delta t_{ij} - \frac{1}{2}\bm{g}\Delta t_{ij}^2 \big)\\
        &= \sum_{k=i}^{j-1} \Big( \Delta \bm{v}_{ik} \Delta t + \frac{1}{2} \Delta \mathtt{R}_{ik} (\tilde{\bm{a}}_k - \bm{b}^a - \bm{\eta}^{ad})\Delta t^2 \Big)
    \end{split}
\end{align}
where we dropped the reference frame subscripts and used $\cdot_t$ instead of $\cdot(t)$ for simplicity of notation. We refer the interested reader to the original paper for more details and derivations.

On the other hand, in a visual-inertial configuration, the sensor poses at the $i$-th keyframe, $\{\mathtt{R}_i, \bm{p}_i\} \in \text{SE(3)}$, can be independently estimated by the vision system. Thus, we have:
\begin{align}
    \mathtt{R}_i &= \bar{\mathtt{R}}_i\,\mathtt{R}_{\rf{CB}}\\
    \bm{p}_i &= s\,\bar{\bm{p}}_i + \mathtt{R}_i\,{}_{\rf{C}}\bm{t}_{\rf{B}}
\end{align}
where $\{\bar{\mathtt{R}}_i, \bar{\bm{p}}_i\} \in \text{SE(3)}$ represent the keyframe poses as estimated by the vision system, $s \in \REAL^+$ the scale factor and $\{\mathtt{R}_{\rf{CB}},{}_{\rf{C}}\bm{t}_{\rf{B}}\} \in \text{SE(3)}$ the extrinsic calibration of the camera with respect to the body frame, which is considered fixed. 

\subsection{Maximum Likelihood Estimation}
The goal of the IMU initialization step is to estimate its initial parameters in order to allow VI fusion. From a MLE perspective, this is:
\begin{equation}
    \label{eq:map}
    \mathcal{X}^\star = \argmax_{\mathcal{X}} \; p(\mathcal{X} \mid \mathcal{Z})
\end{equation}
where $\mathcal{X}$ is the set of variables we want to estimate and $\mathcal{Z}$ represents the set of observations. Under the mild assumption of independence between observations, we can write the posterior probability as:
\begin{equation}
    p(\mathcal{X} \mid \mathcal{Z}) \propto p(\mathcal{Z} \mid \mathcal{X}) = \prod_{k \in \mathcal{K}} p(\mathcal{Z}_k \mid \mathcal{X})
\end{equation}
for each independent observation $k$. Under the assumption of zero-mean Gaussian noise, the optimization problem in~\eqref{eq:map} can be written as the sum of the squared residual errors:
\begin{equation}
    \label{eq:optimization_problem}
    \mathcal{X}^\star = \argmin_{\mathcal{X}} \; \sum_{k \in \mathcal{K}} \norm{\bm{r}_k}^2_{\mathbf{\Sigma}_k},
\end{equation}
where $\bm{r}_k$ are the residual errors and $\mathbf{\Sigma}_k$ their corresponding covariance matrices.

In the following section, we express these MLE-based equations as a function of the parameters to be estimated.

\section{IMU Initialization}
The complete VI-system initialization process, as proposed in~\cite{campos2020}, can be summarized as follows:
\begin{enumerate}
    \item \emph{Vision-only estimation}. This step usually involves the initialization of the vision system followed by a bundle-adjustment process for a few seconds of data, producing an up-to-scale estimate of the keyframe poses.
    \item \emph{IMU initialization}. Provides an initial estimate for the IMU-related parameters, as well as for the scale factor, from the previously estimated keyframe poses.
    \item \emph{Visual-inertial refinement}. Optionally, all previous parameters estimated separately for the vision and inertial systems can be refined in a combined VI-BA.
\end{enumerate}

The first step can be addressed through any visual-based odometry/SLAM system such as ORB-SLAM~\cite{orb_slam} or PL-SLAM~\cite{pl_slam}, as long as it provides accurate enough keyframe poses estimations, while the third, optional, stage can be accomplished by means of a standard VI bundle-adjustment procedure that considers both visual features and inertial measurements~\cite{mur2017}.

Is in the second step of the process where our method contributes with an analytic solution for the estimation of the IMU parameters (except the gyroscope bias), as formally stated next. For that, the IMU initialization process is further divided into two subproblems: (i) iteratively estimating the gyroscope bias and (ii) solving for the rest of the parameters. Our approach relies on the mild assumptions that the vision-only poses are very accurate (up to a scale factor), and that the IMU biases can be considered constant during the initialization.

\subsection{Gyroscope bias}
First we deal with the estimation of the gyroscope bias, that is, $\mathcal{X}_g\ \dot{=}\ \{ \bm{b}^g \}$, from the gyroscope measurements and the keyframe orientations, i.e., $\mathcal{Z}_k\ \dot{=}\ \{ \Delta \mathtt{R}_{k,k+1}, \bar{\mathtt{R}}_{k:k+1} \}$. For that, we define the residual function from the preintegrated rotation \eqref{eq:preintegrated_rotation} as:
\begin{gather}
    \bm{r}_k^g(\bm{x})\ \dot{=}\ \Log\Big(\big(\Delta \mathtt{R}_{k,k+1}\Exp(\mathbf{J}_{\Delta\mathtt{R}}^g\bm{x})\big)^\top\,\mathtt{R}_{k}^\top\,\mathtt{R}_{k+1}\Big),
\end{gather}
where $\Log(\cdot)$ stands for the inverse of the SO(3) exponential map and $\mathbf{J}_{\Delta\mathtt{R}}^g$ represents the Jacobian of the preintegrated rotation with respect to the bias.

Therefore, the MLE estimation of the gyroscope bias is equivalent to this optimization problem:
\begin{equation}
    \label{eq:gyro_optimization}
    \mathcal{X}_g^\star = \argmin_{\bm{b}^g} \; \sum_{k \in \mathcal{K}} \norm{\bm{r}_k^g(\bm{b}^g)}^2_{\mathbf{\Sigma}^g_k}
\end{equation}
which can be solved iteratively using the well-known Levenberg–Marquardt algorithm starting from a zero bias. The analytic expressions for the preintegrated Jacobians and uncertainty propagation can be found in~\cite{forster2017}.

\subsection{Accelerometer bias, gravity direction and scale factor}
For the accelerometer, in turn, we are interested in finding analytically its bias ($\bm{b}^a$), the gravity direction ($\bm{g}$) and the scale ($s$) of the reconstruction from the measured acceleration and the relative pose.
We opt here to eliminate the velocities by considering three consecutive keyframes in order to reduce the complexity of the problem, as proposed in~\cite{mur2017}. Therefore, we have:
\begin{gather}
    \label{eq:acc_variables}
    \mathcal{X}_a\ \dot{=}\ \{ s, \bm{b}^a, \bm{g} \}\\
    \label{eq:acc_observations}
    \mathcal{Z}_k\ \dot{=}\ \{ \Delta \bm{v}_{k-1:k+1}, \Delta \bm{p}_{k-1:k+1}, \bm{b}^g, \bar{\mathtt{R}}_{k-1:k+1}, \bar{\bm{p}}_{k-1:k+1} \}
\end{gather}

Thus, from \eqref{eq:preintegrated_velocity} and \eqref{eq:preintegrated_position}, we define the residual for the accelerometer as:
\begin{gather}
    \bm{r}_k^a(\bm{x}) =
    \begin{bmatrix}
        \bm{\alpha}_k & \mathbf{A}_k & \mathbf{B}_k
    \end{bmatrix} \bm{x} - \bm{\pi}_k
\end{gather}
with
\begin{gather}
    \mathbf{A}_k = \frac{\mathtt{R}_{k-1}\mathbf{J}_{\Delta \bm{p}_{k-1,k}}^a}{\Delta t_{k-1,k}} - \frac{\mathtt{R}_{k}\mathbf{J}_{\Delta \bm{p}_{k,k+1}}^a}{\Delta t_{k,k+1}} - \mathtt{R}_{k-1}\mathbf{J}_{\Delta \bm{v}_{k-1,k}}^a\\
    \mathbf{B}_k = -\frac{1}{2}(\Delta t_{k-1,k} + \Delta t_{k,k+1})\mathbf{I}_{3 \times 3}\\
    \bm{\alpha}_k = \frac{\bar{\bm{p}}_{k+1}-\bar{\bm{p}}_{k}}{\Delta t_{k,k+1}} - \frac{\bar{\bm{p}}_{k}-\bar{\bm{p}}_{k-1}}{\Delta t_{k-1,k}}, \quad \bm{x} = \begin{bmatrix} s & \bm{b}^a & \bm{g} \end{bmatrix}^\top
\end{gather}
\begin{align}
    \bm{\pi}_k =& \frac{\mathtt{R}_{k}\Delta \bm{p}_{k,k+1}}{\Delta t_{k,k+1}} - \frac{\mathtt{R}_{k-1}\Delta \bm{p}_{k-1,k}}{\Delta t_{k-1,k}} + \mathtt{R}_{k-1}\Delta \bm{v}_{k-1,k}\\ \nonumber
    & + \frac{\big(\bar{\mathtt{R}}_{k} - \bar{\mathtt{R}}_{k-1}\big){}_{\rf{C}}\bm{t}_{\rf{B}}}{\Delta t_{k-1,k}} - \frac{\big(\bar{\mathtt{R}}_{k+1} - \bar{\mathtt{R}}_{k}\big)}{\Delta t_{k,k+1}},
\end{align}
where $\mathbf{J}_{\Delta \bm{v}_{ij}}^a$ and $\mathbf{J}_{\Delta \bm{p}_{ij}}^a$ represent the Jacobian matrices of the preintegrated velocity and position with respect to the accelerometer bias, respectively. As before, the analytic expressions for these Jacobians, as wel as their derivations, can be found in~\cite{forster2017}.

To sum up, the residual function comes from taking the difference $\nicefrac{(\bm{p}_{k+1} - \bm{p}_k)}{\Delta t_{k,k+1}} - \nicefrac{(\bm{p}_k - \bm{p}_{k-1})}{\Delta t_{k-1,k}}$ from the preintegrated position~\eqref{eq:preintegrated_position} and substituting the expression $\bm{v}_{k+1} - \bm{v}_k$ involving the velocities with the preintegrated velocity from~\eqref{eq:preintegrated_velocity}.

Note that, in this case, we are facing a constrained optimization problem, since the magnitude of the gravity vector is known:
\begin{align}
    \label{eq:acc_optimization}
    \begin{split}
        \mathcal{X}_a^\star = &\argmin_{\bm{x}} \; \sum_{k \in \mathcal{K}} \norm{\bm{r}_k^a(\bm{x})}^2_{\mathbf{\Sigma}^a_k}\\
        &\text{subject to}\; \norm{\bm{g}} = G,
    \end{split}
\end{align}
where $G = 9.81$ is the magnitude of the gravity vector.

To solve this problem, first we write the cost function \eqref{eq:acc_optimization} in matrix form (where the superscript $\cdot^a$ in $\Sigma^a_k$ has been omitted for clarity):
\begin{align}
    \label{eq:acc_cost_matrix}
    \begin{split}
        \mathcal{C}(\bm{x}) &= \sum_{k \in \mathcal{K}} \norm{\bm{r}_k^a(\bm{x})}^2_{\mathbf{\Sigma}_k}\\
        &= \sum_{k \in \mathcal{K}} (\mathbf{M}_k \bm{x} - \bm{\pi}_k)^\top \mathbf{\Sigma}^{-1}_k (\mathbf{M}_k \bm{x} - \bm{\pi}_k)\\
        &= \sum_{k \in \mathcal{K}} \big( \bm{x}^\top \mathbf{M}^\top_k \Sigma^{-1}_k \mathbf{M}_k \bm{x} - 2\bm{\pi}_k^\top \Sigma^{-1}_k \mathbf{M}_k \bm{x} + \bm{\pi}_k^\top \Sigma^{-1}_k \bm{\pi} \big)\\
        &= \bm{x}^\top \mathbf{M} \bm{x} + \bm{m}^\top \bm{x} + Q,
    \end{split}
\end{align}
where
\begin{equation}
    \begin{split}
        \mathbf{M} = \sum_{k \in \mathcal{K}} \mathbf{M}^\top_k \Sigma^{-1}_{k}\mathbf{M}_k,\quad \bm{m}^\top = \sum_{k \in \mathcal{K}} -2\bm{\pi}_k^\top \Sigma^{-1}_k \mathbf{M}_k,\\
        Q = \sum_{k \in \mathcal{K}} \bm{\pi}^\top_k \Sigma^{-1}_k \bm{\pi}_k,\quad \mathbf{M}_k = \begin{bmatrix}
        \bm{\alpha}_k & \mathbf{A}_k & \mathbf{B}_k
    \end{bmatrix}
    \end{split}
\end{equation}

We also write the constraint \eqref{eq:acc_optimization} in quadratic form:
\begin{equation}
    \label{eq:acc_quadratic_constraint}
    \bm{x}^\top \mathbf{W} \bm{x} = G^2,\quad
    \mathbf{W} =
    \begin{bmatrix}
        \mathbf{0}_{4 \times 4} & \bm{0}^\top\\
        \bm{0} & \mathbf{I}_{3 \times 3}
    \end{bmatrix},
\end{equation}
so that the problem in \eqref{eq:acc_optimization} can be expressed as:
\begin{equation}
    \label{eq:acc_quadratic_system}
    \begin{split}
        \mathcal{X}_a^\star = \argmin_{\bm{x}}\ \bm{x}^\top \mathbf{M} \bm{x} + \bm{m}^\top \bm{x}\\
        \text{subject to}\ \bm{x}^\top \mathbf{W} \bm{x} = G^2
    \end{split}
\end{equation}

Then, we solve this quadratic system with a quadratic constraint using Lagrange's multipliers. For that, let the Lagrangian be:
\begin{equation}
    \mathcal{L}(\bm{x}) = \bm{x}^\top \mathbf{M} \bm{x} + \bm{m}^\top \bm{x} + \lambda(\bm{x}^\top \mathbf{W} \bm{x} - G^2)
\end{equation}

Therefore, the necessary condition for optimality is:
\begin{equation}
    \frac{\partial \mathcal{L}(\bm{x})}{\partial \bm{x}} = 2\bm{x}^\top \mathbf{M} + \bm{m}^\top + 2\lambda\bm{x}^\top \mathbf{W} = \bm{0}^\top,
\end{equation}
or, equivalently:
\begin{equation}
    \label{eq:acc_quadratic_solution}
    \bm{x} = - (2\mathbf{M} + 2\lambda\mathbf{W})^{-\top}\bm{m}.
\end{equation}

Finally, substituting the relation in \eqref{eq:acc_quadratic_solution} back to the constraint in \eqref{eq:acc_quadratic_constraint}, leads to:
\begin{equation}
    \label{eq:acc_lambda_constraint}
    \bm{m}^\top (2\mathbf{M} + 2\lambda\mathbf{W})^{-1} \mathbf{W} (2\mathbf{M} + 2\lambda\mathbf{W})^{-\top}\bm{m} = G^2
\end{equation}
where $\lambda$ is now the only unknown. It can be shown that \eqref{eq:acc_lambda_constraint} is a sixth-order polynomial in $\lambda$ (see the appendix for a proof, cf.~\cite{censi2008}). Thereby, we reduce the above-defined constrained optimization problem to finding the roots of this polynomial, which can be solved, for instance, by computing the eigenvalues of the corresponding companion matrix~\cite{matrix_analysis}. Once solved for $\lambda$, the original solution can be recovered using \eqref{eq:acc_quadratic_solution}. 

This analytical procedure does not need any initial guess for the parameters to be estimated and, as proven in our experiments, takes about $7\times$ less time to get the solution than previous iterative approaches, while achieving similar accuracy.

\section{Experiments}
We validated and evaluated the proposed IMU initialization approach with real data from the EuRoC dataset~\cite{euroc}. We chose this dataset since it provides accurate groundtruth from a Vicon motion capture system along the whole trajectory, for a total of \num{11} sequences. In the experiments, an IMU initialization procedure is attempted every \SI{0.5}{\second}, with keyframes inserted at a fixed rate of \SI{4}{\hertz}. The results are compared with the state-of-the-art IMU initialization approach described in~\cite{campos2020} for reference (denoted by \emph{inertial-only} in the following). In turn, the iterative optimization problems are implemented using Ceres Solver~\cite{ceres-solver}, while we used Eigen for our proposed analytical solution\footnote{The reference implementation of the proposed IMU initialization approach is publicly available at: \url{https://github.com/dzunigan/imu_initialization}}. All experiments were executed in an Intel Core i7-7700HQ CPU with \SI{16}{\giga\byte} of RAM.

\subsection{Validation}
In this experiment we executed both, our method and the \emph{inertial-only} optimizatio, using the keyframe poses from the dataset groundtruth instead of the vision-only estimated ones, in order to serve as a baseline of the expected performance.
The estimated IMU parameters, i.e. biases, gravity direction and scale factor, are then compared with the groundtruth parameters provided by the dataset. The results of this experiment are shown in \tabref{tab:validation_results}. Note that, for the gravity vector, only the angle is considered for evaluation, while, for the biases, only the magnitude is considered. Besides, attempts of initialization with near constant velocity are unobservable~\cite{martinelli2014}, and thus are discarded. Because of this, and in order to get a fair comparison, we use the same heuristic proposed in~\cite{campos2020}, which discards attempts that exhibit small accelerations (average acceleration within \SI{0.5}{\percent} of the gravity magnitude).

\tabvalidation

As shown in the results, our approach performs better in general for the scale factor, even though the true scale factor is used as the initial guess for the \emph{inertial-only} optimization. In turn, the gyroscope bias is solved in both cases using the same iterative algorithm, but we achieve slightly better results since we solve it separately from the rest of the parameters, therefore dealing with a reduced problem. For the accelerometer bias and the direction of the gravity vector, the \textit{inertial-only} approach produces better results for short times of data (about \SI{5}{\second}). However our approach achieves higher accuracy when considering longer initialization times, presenting errors with a clear diminishing tendency, in contrast to~\cite{campos2020}, which decrease at a slower rate.

The particular case of the accelerometer bias deserves more attention, though, as it is well-known that such bias is hard to estimate for short initialization times (see~\cite{mur2017}, for example). To deal with this, Campos \etal~\cite{campos2020} proposed to introduce a prior residual for this bias in a MAP formulation. For this comparison, we set it to a fixed value of \num{e5} (as in the implementation of ORB\_SLAM3~\cite{campos2020orb} system by the same authors), so that the bias remains close to zero for the whole experiment. In contrast, our method does not rely on any prior value, and still is able to estimate the bias more accurately than~\cite{campos2020} when using more than \SI{5}{\second} of data.

\begin{figure}[t]
    \centering
    \includegraphics[width=0.47\textwidth]{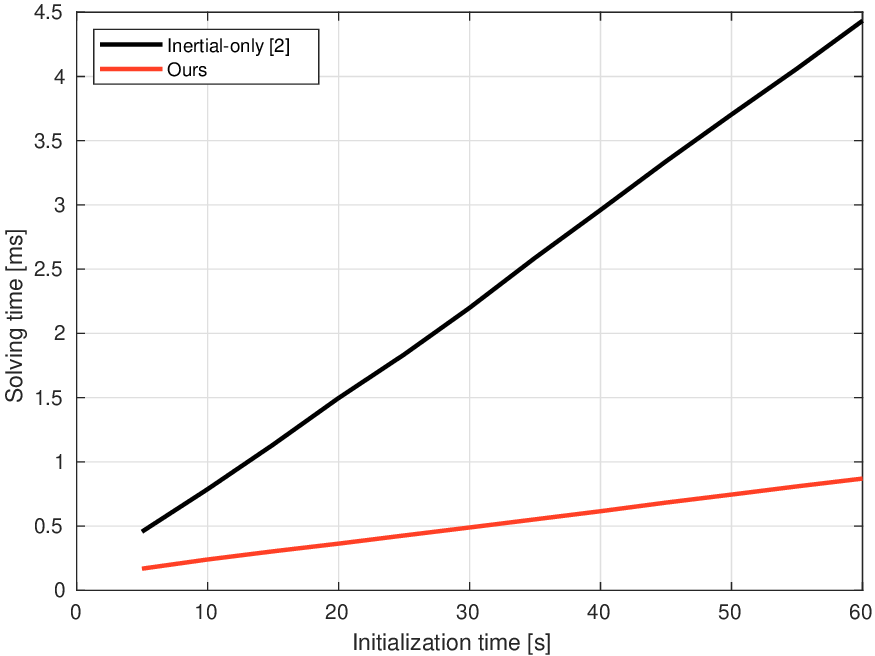}
    \caption{Required solving time (in average for all dataset sequences) with respect to the length of the considered interval for initialization.}
    \label{fig:solving_time}
\end{figure}

Finally, in \figref{fig:solving_time} we show the required time for solving the IMU initialization (average of all the dataset sequences) as a function of the time considered for initialization. Note that, for this test, we set again the initial guess for the metric scale to its true value for the \emph{inertial-only} approach. As can be seen, even in these conditions our solution is faster and scales better compared to the iterative approach in~\cite{campos2020}.

\subsection{Evaluation}
In this experiment, we used the output of a visual odometry system, instead of the groundtruth, in order to perform the IMU initialization. We chose the well-known open source ORB\_SLAM2~\cite{orb_slam2} system for this purpose, processing the images in monocular mode from the left camera only. We also disabled the loop closure and keyframe culling and run ORB\_SLAM2 once for each sequence in the dataset. If the tracking was lost, we run the system again from were it failed. As in the previous experiment, we rejected initializations with small accelerations and also used a fixed keyframe insertion rate of \SI{4}{\hertz}. We also tested a \emph{standard} non-linear optimization alternative, that is, the same approach than~\cite{campos2020} but without any bias prior. In \tabref{tab:evaluation_results} we show the evaluation results of the IMU initialization process considering \num{20} keyframes (\SI{5}{\second}).

\tabevaluation

In this case, our approach produced slightly better estimations for the scale factor and the gyroscope bias. However, with only \num{20} keyframes, the accelerometer bias could not be reliably estimated, consequently affecting the estimation of the gravity direction. 
In this experiment, we used the three initial guesses for the scale factor proposed in the original paper~\cite{campos2020}: \num{1}, \num{4}, \num{16}, working in parallel.
It is worth noting that we only show the required solving time, i.e. without taking into account the IMU preintegration, which is common for all methods (taking about \SI{3.5}{\milli\second} per keyframe, for a \SI{200}{\hertz} IMU and keyframes captured at \SI{4}{\hertz}). The solution proposed by Campos \etal\ is already very efficient, requiring about \SI{1.3}{\milli\second} on average to solve the initialization parameters. Comparatively, though, the solution presented in this paper runs about $\times7$ faster in average.

\section{Conclusion}
In this work we have addressed the initialization of the IMU parameters in a monocular Visual-Inertial system and proposed an analytical solution for the estimation of the accelerometer bias, the direction of gravity and the scale factor of the vision-based reconstruction.

The proposed formulation is defined within a MLE framework and derives a sixth degree polynomial that can be efficiently solved to obtain optimal values for such parameters. This way, our approach avoids the intrinsic issues of previous iterative solutions, including the need for a proper initial guess of the parameter values, specially the scale factor, and the possibility of divergence of the procedure.
Additionally, our implementation incurs in only a fraction of the computational burden reported by state-of-the-art methods, being about \num{7} times faster.

Our solution has been validated and extensively assessed in a series of experiments with real data from the popular visual-inertial EuRoC dataset and using the vision-based ORB\_SLAM2~\cite{orb_slam2} system, yielding consistent results in all cases. Nevertheless, as the experimental evaluation suggests, the observability of the accelerometer bias should be explicitly addressed in future work.
Finally, our implementation of the IMU initialization proposed in this work can be found at: \url{https://github.com/dzunigan/imu_initialization}.

\bibliographystyle{IEEEtran}
\bibliography{ref}

\appendix
\label{appendix}

\emph{Proposition:} Equation \eqref{eq:acc_lambda_constraint} is a sixth-order polynomial in $\lambda$.

\noindent \emph{Proof:} Partition the matrix $(2\mathbf{M} + 2\lambda\mathbf{W})$ into four sub-matrices:
\begin{equation}
    2\mathbf{M} + 2\lambda\mathbf{W} =
    \begin{bmatrix}
    \mathbf{A} & \mathbf{B}\\
    \mathbf{B}^\top & \mathbf{D} + 2\lambda\mathbf{I}_{3 \times 3}
    \end{bmatrix}
\end{equation}
Since the $\mathbf{W}$ in the quadratic form \eqref{eq:acc_lambda_constraint} is sparse, we only need to compute the last column of $(2\mathbf{M} + 2\lambda\mathbf{W})^{-1}$. Using the matrix inversion lemma we have:
\begin{equation}
    \begin{bmatrix}
    \mathbf{A} & \mathbf{B}\\
    \mathbf{B}^\top & \mathbf{D} + 2\lambda\mathbf{I}_{3 \times 3}
    \end{bmatrix}^{-1} =
    \begin{bmatrix}
    \star & -\mathbf{A}^{-1}\mathbf{B}\mathbf{Q}^{-1}\\
    \star & \mathbf{Q}^{-1}
    \end{bmatrix}
\end{equation}
where $\mathbf{Q} = \mathbf{D} - \mathbf{B}^\top \mathbf{A}^{-1}\mathbf{B} + 2\lambda\mathbf{I}_{3 \times 3}\ \dot{=}\ \mathbf{S} + 2\lambda\mathbf{I}_{3 \times 3}$

The quadratic constraint \eqref{eq:acc_lambda_constraint} can now be expressed as:
\begin{gather}
    \label{eq:qcc_compact_matrix}
    \bm{m}^\top \mathbf{F} \bm{m} = G^2\\
    \mathbf{F} =
    \begin{bmatrix}
        \mathbf{A}^{-1}\mathbf{B}\mathbf{Q}^{-1}\mathbf{Q}^{-\top}\mathbf{B}^\top\mathbf{A}^{-\top} & -\mathbf{A}^{-1}\mathbf{B}\mathbf{Q}^{-1}\mathbf{Q}^{-\top}\\
        \text{symm} & \mathbf{Q}^{-1}\mathbf{Q}^{-\top}
    \end{bmatrix}
\end{gather}
where only the matrix $\mathbf{Q}^{-1}$ depends on $\lambda$. The analytic expression for this matrix can be written as:
\begin{align}
    \mathbf{Q}^{-1} &= (\mathbf{S} + 2\lambda\mathbf{I}_{3 \times 3})^{-1} = \frac{(\mathbf{S} + 2\lambda\mathbf{I}_{3 \times 3})^{\mathrm{A}}}{p(\lambda)}\\
    &= \frac{\mathbf{S}^{\mathrm{A}} + 4\lambda^2\mathbf{I}_{3 \times 3} + 2\lambda\big(\tr(\mathbf{S})\mathbf{I}_{3 \times 3} - \mathbf{S}\big)}{p(\lambda)}
\end{align}
where
\begin{equation}
    p(\lambda)\ \dot{=}\ \det(\mathbf{S} + 2\lambda\mathbf{I}_{3 \times 3}),\quad \mathbf{S}^{\mathrm{A}} = \det(\mathbf{S}) \mathbf{S}^{-1}
\end{equation}

Since $\mathbf{Q}^{-1}$ is symmetric, we have:
\begin{align}
    \mathbf{Q}^{-1}\mathbf{Q}^{-\top} &= \frac{(\mathbf{S}^{\mathrm{A}} + 4\lambda^2\mathbf{I}_{3 \times 3} + 2\lambda\mathbf{U})^2}{p(\lambda)^2}\\
    \begin{split}
        &= \frac{16\lambda^4\mathbf{I}_{3 \times 3} + 4\lambda^2(2\mathbf{S}^{\mathrm{A}} + \mathbf{U}^2)}{p(\lambda)^2}\\
        &+ \frac{16\lambda^3\mathbf{U} + 2\lambda(\mathbf{S}^{\mathrm{A}}\mathbf{U} + \mathbf{U}\mathbf{S}^{\mathrm{A}}) + {\mathbf{S}^{\mathrm{A}}}^2}{p(\lambda)^2}
    \end{split}
\end{align}
where
\begin{equation}
    \mathbf{U}\ \dot{=}\ \tr(\mathbf{S})\mathbf{I}_{3 \times 3} - \mathbf{S}
\end{equation}

Therefore, \eqref{eq:acc_lambda_constraint} can be written as a polynomial expression on $\lambda$:
\begin{equation}
    \label{eq:acc_polynomial}
    \begin{split}
        16\lambda^4\ \bm{m}^\top
        \begin{bmatrix}
            \mathbf{A}^{-1}\mathbf{B}\mathbf{B}^\top\mathbf{A}^{-\top} & -\mathbf{A}^{-1}\mathbf{B}\\
            \text{symm} & \mathbf{I}_{3 \times 3}
        \end{bmatrix}
        \bm{m}\\
        + 16\lambda^3\ \bm{m}^\top
        \begin{bmatrix}
            \mathbf{A}^{-1}\mathbf{B}\mathbf{U}\mathbf{B}^\top\mathbf{A}^{-\top} & -\mathbf{A}^{-1}\mathbf{B}\mathbf{U}\\
            \text{symm} & \mathbf{U}
        \end{bmatrix}
        \bm{m}\\
        + 4\lambda^2\ \bm{m}^\top
        \begin{bmatrix}
            \mathbf{A}^{-1}\mathbf{B}\mathbf{X}\mathbf{B}^\top\mathbf{A}^{-\top} & -\mathbf{A}^{-1}\mathbf{B}\mathbf{X}\\
            \text{symm} & \mathbf{X}
        \end{bmatrix}
        \bm{m}\\
        + 2\lambda\ \bm{m}^\top
        \begin{bmatrix}
            \mathbf{A}^{-1}\mathbf{B}\mathbf{Y}\mathbf{B}^\top\mathbf{A}^{-\top} & -\mathbf{A}^{-1}\mathbf{B}\mathbf{Y}\\
            \text{symm} & \mathbf{Y}
        \end{bmatrix}
        \bm{m}\\
        + \bm{m}^\top
        \begin{bmatrix}
            \mathbf{A}^{-1}\mathbf{B}{\mathbf{S}^{\mathrm{A}}}^2\mathbf{B}^\top\mathbf{A}^{-\top} & -\mathbf{A}^{-1}\mathbf{B}{\mathbf{S}^{\mathrm{A}}}^2\\
            \text{symm} & {\mathbf{S}^{\mathrm{A}}}^2
        \end{bmatrix}
        \bm{m}\\
        = G^2p(\lambda)^2
    \end{split}
\end{equation}
where
\begin{gather}
    \mathbf{X}\ \dot{=}\ 2\mathbf{S}^{\mathrm{A}} + \mathbf{U}^2,\quad \mathbf{Y}\ \dot{=}\ \mathbf{S}^{\mathrm{A}}\mathbf{U} + \mathbf{U}\mathbf{S}^{\mathrm{A}}
\end{gather}

Since $p(\lambda)$ is a cubic function of $\lambda$, the expression \eqref{eq:acc_polynomial} is a sixth-order polynomial in $\lambda$.

\end{document}

%% file: tables.tex
\newcommand\tstrut{\rule{0pt}{2.6ex}}       
\newcommand\bstrut{\rule[-1.5ex]{0pt}{0pt}} 

\newcommand\tabvalidation{
\begin{table}[tb]
\begin{center}
    \small\addtolength{\tabcolsep}{-4pt}
    \caption{Errors of the estimated parameters from groundtruth poses}
    \label{tab:validation_results}
    \scriptsize
    \begin{tabular}{|c||c c c c | c c c c||}
    \hline
    
    \multirow{2}{*}{Time} & \multicolumn{4}{c|}{\tstrut Inertial-only Optimization~\cite{campos2020}} & \multicolumn{4}{c||}{Proposed IMU Initialization\bstrut} \\ \cline{2-9}
    & Scale & Gyro Bias & Acc Bias & G. Dir & Scale & Gyro Bias & Acc Bias & \tstrut G. Dir\bstrut \\ \hline \hline
    
     1.25s & 6.23\% & 1.16\% & 99.9\% & 1.11º & 4.61\% & 1.16\% & 721\% & \tstrut 7.6º \\
    
     2.5s  & 3.18\% & 0.98\% & 99.9\% & 1.06º & 2.57\% & 0.94\% & 299\% & 3.24º \\
    
     5s    & 2.64\% & 0.83\% & 99.7\% & 1.02º & 1.60\% & 0.76\% & 90.3\% & 1.18º \\
    
    12.5s  & 2.30\% & 0.72\% & 99.1\% & 0.94º & 1.21\% & 0.52\% & 21.6\% & 0.42º \\
    
    18.75s & 2.05\% & 0.47\% & 98.5\% & 0.88º & 1.11\% & 0.35\% & 12.7\% & 0.29º\bstrut \\ \hline
    
    \end{tabular}
\end{center}
\end{table}
}

\newcommand\tabevaluation{
\begin{table*}[tb]
\begin{center}
    \small\addtolength{\tabcolsep}{-4pt}
    \caption{Initialization times and errors of the estimated parameters form visual poses}
    \label{tab:evaluation_results}
    \scriptsize
    \begin{tabular}{|c||c c c c c | c c c c c | c c c c c||}
    \hline
    
    \multirow{2}{*}{Sequence} & \multicolumn{5}{c|}{\tstrut Inertial-only Optimization~\cite{campos2020}} & \multicolumn{5}{c|}{Standard} & \multicolumn{5}{c||}{Proposed IMU Initialization\bstrut} \\ \cline{2-16}
    & Solve & Scale & Gyro Bias & Acc Bias & G. Dir & Solve & Scale & Gyro Bias & Acc Bias & G. Dir & Solve & Scale & Gyro Bias & Acc Bias & \tstrut G. Dir\bstrut \\ \hline \hline
    
    V1\_01 & 1.34 ms & 27.9\% & 0.72\% & 99.9\% & 2.85º & 1.09 ms & 29.2\% & 0.86\% & 56.3\% & 3.04º & 0.18 ms & 19.5\% & 0.84\% & 58.5\% & \tstrut 2.93º \\
    
    V1\_02 & 1.05 ms & 4.50\% & 0.76\% & 99.6\% & 1.01º & 1.09 ms & 4.15\% & 1.22\% & 43.2\% & 0.88º & 0.18 ms & 2.69\% & 1.27\% & 33.4\% & 0.81º \\
    
    V1\_03 & 0.90 ms & 13.5\% & 2.72\% & 99.6\% & 2.55º & 1.05 ms & 15.1\% & 2.90\% & 261\%  & 5.12º & 0.18 ms & 11.3\% & 2.48\% & 203\% & 4.54º \\
    
    V2\_01 & 1.09 ms & 8.66\% & 0.76\% & 99.7\% & 2.16º & 1.04 ms & 8.78\% & 0.79\% & 126\%  & 2.97º & 0.19 ms & 6.36\% & 0.76\% & 96.5\% & 2.72º \\
    
    V2\_02 & 1.02 ms & 6.54\% & 0.99\% & 99.6\% & 1.49º & 1.11 ms & 7.18\% & 0.90\% & 327\%  & 2.84º & 0.18 ms & 5.90\% & 0.75\% & 125\% & 1.55º \\
    
    V2\_03 & 0.93 ms & 4.90\% & 0.77\% & 99.6\% & 1.53º & 1.04 ms & 4.66\% & 0.60\% & 118\%  & 1.59º & 0.19 ms & 2.67\% & 0.55\% & 98.2\% & 1.55º \\
    
    MH\_01 & 1.32 ms & 26.4\% & 0.83\% & 99.7\% & 1.40º & 2.97 ms & 25.4\% & 0.83\% & 475\%  & 7.15º & 0.19 ms & 15.3\% & 0.71\% & 371\% & 4.68º \\
    
    MH\_02 & 0.99 ms & 12.1\% & 0.40\% & 99.7\% & 1.48º & 1.01 ms & 12.5\% & 0.41\% & 269\%  & 3.73º & 0.19 ms & 10.0\% & 0.40\% & 228\% & 3.26º \\
    
    MH\_03 & 1.19 ms & 18.2\% & 0.60\% & 99.7\% & 1.39º & 1.12 ms & 20.4\% & 0.57\% & 531\%  & 5.73º & 0.19 ms & 14.2\% & 0.50\% & 299\% & 3.39º \\
    
    MH\_04 & 2.23 ms & 17.8\% & 0.95\% & 99.7\% & 1.52º & 2.15 ms & 19.5\% & 0.93\% & 736\%  & 7.23º & 0.19 ms & 15.1\% & 0.87\% & 585\% & 5.85º \\
    
    MH\_05 & 2.17 ms & 19.8\% & 1.21\% & 99.7\% & 6.44º & 1.72 ms & 20.8\% & 1.20\% & 870\%  & 11.4º & 0.18 ms & 11.7\% & 1.01\% & 439\% & 5.15º\bstrut\\ \hline \hline
    
    Avg.   & 1.29 ms & 14.6\% & 0.97\% & 99.7\% & 2.17º & 1.40 ms & 15.2\% & 1.02\% & 330\%  & 4.70º & 0.19 ms & 10.4\% & 0.92\% & 230\% & \tstrut 3.44º\bstrut \\ \hline
    
    \end{tabular}
\end{center}
\end{table*}
}